\documentclass[11pt,a4paper]{article}
\usepackage{times,latexsym}
\usepackage{url}

\usepackage[acceptedWithA]{tacl2021v1}
\usepackage[T1]{fontenc}

\usepackage{xspace,mfirstuc,tabulary}

\newif\iftaclinstructions
\taclinstructionsfalse 
\iftaclinstructions
\renewcommand{\confidential}{}
\renewcommand{\anonsubtext}{(No author info supplied here, for consistency with
TACL-submission anonymization requirements)}
\newcommand{\instr}
\fi

\iftaclpubformat 

\else

\fi

\newcommand{\datasetnamenospace}{\textsc{TyDi~QA--WANA}}

\usepackage[most]{tcolorbox}
\newtcolorbox{examplebox}[1][]{
  enhanced,
  attach boxed title to top left={yshift=-2mm, xshift=2mm}, 
  boxrule=1pt,                  
  colback=gray!15,              
  colframe=black,               
  colbacktitle=black,           
  coltitle=white,               
  fonttitle=\sffamily\bfseries, 
  sharp corners,                
  title=#1
}
\usepackage{enumitem}
\usepackage[utf8]{inputenc}
\usepackage{tipa}
\usepackage{newunicodechar}
\newunicodechar{ə}{\textschwa}
\usepackage{bbding}
\newcommand{\githublink}{https://github.com/google-research-datasets/tydiqa-wana}

\title{{\textbf{\scshape TyDi~QA--WANA}}: A Benchmark for Information-Seeking Question Answering
in Languages of \underline{W}est \underline{A}sia and \underline{N}orth \underline{A}frica}

\author{
  Parker Riley
  \\
  Google
  \And
  Siamak Shakeri
  \\
  Google
  \And
  Waleed Ammar\Thanks{Work done at Google.}
  \\
  Holistic Intelligence for 
  \\
  Global Good
  \And
  Jonathan H. Clark
  \\
  Google
}

\date{}

\begin{document}
\maketitle
\begin{abstract}
We present \datasetnamenospace, a question-answering dataset consisting of 28K examples divided among 10 language varieties of western Asia and northern Africa. The data collection process was designed to elicit \textit{information-seeking} questions, where the asker is genuinely curious to know the answer. Each question in paired with an entire article that may or may not contain the answer; the relatively large size of the articles results in a task suitable for evaluating models' abilities to utilize large text contexts in answering questions.
Furthermore, the data was collected directly in each language variety, without the use of translation, in order to avoid issues of cultural relevance.
We present performance of two baseline models, and release our code and data to facilitate further improvement by the research community.
\end{abstract}
\section{Introduction}

Many users worldwide regularly use technology to help them answer \textit{information-seeking} questions, where the user does not know the answer but wants to. When developing systems to answer these questions, improving quality requires being able to reliably measure it using trustworthy and challenging evaluations. For measuring performance on answering English questions, multiple evaluation datasets are available. Some datasets are also available in non-English language varieties, but they cover only a tiny portion of the language varieties of the world.

Modern LLMs have shown promising performance in QA tasks, including non-English ones \citep{geminiteam2024geminifamilyhighlycapable}. However, performance is known to be worse for low-resource language varieties for which these models have seen comparatively little pretraining data. This highlights the need for training and evaluation data in these languages.

Additionally, the advent of long-context LLMs, which can accept more than 1 million input tokens \cite{reid2024gemini}, poses a challenge in evaluation because most information-seeking datasets are not designed to test a model's ability to utilize such a large context window.

We seek to address these issues by creating and releasing\footnote{Our code and data are available at \githublink.} a dataset of long-context information-seeking questions in under-represented non-English language varieties of West Asia and North Africa, in the style of \textsc{TyDi~QA} \citep{clark-etal-2020-tydi}, which we call \datasetnamenospace. We also present baseline results that illustrate that modern LLMs are capable of answering questions by including \textit{an entire Wikipedia article} in the input, something made feasible by recent advances in long-context modeling.

The rest of this work is organized as follows. First, we define the long-context information-seeking question answering task (\S\ref{sec:task_definition}). Next, we describe how we created the data (\S\ref{sec:data_collection}). We then outline the gap in prior work that this work addresses (\S\ref{sec:related_work}) and catalog the language varieties selected for inclusion in our dataset (\S\ref{sec:language_varieties}). To quantify these contributions, we present basic statistics of our dataset (\S\ref{sec:dataset_statistics}), describe an evaluation procedure (\S\ref{sec:evaluation}), and present results on baseline systems (\S\ref{sec:results}).
\section{Task Definition}\label{sec:task_definition}
In this work we adopt a single task from \citet{clark-etal-2020-tydi}: the \textbf{Minimal Answer Span Task} (abbreviated as MinSpan). In this task, the input is the full text of an article paired with a question, and the desired output is one of three types:
\begin{enumerate}
    \item The start and end byte indices of the minimal span that completely answers the question, if such a span exists within the article.
    \item \textsc{YES} or \textsc{NO} if the question requires a yes/no answer and such a conclusion can be drawn from the article.
    \item \textsc{NULL} if it is not possible to answer the question with one of the above two types.
\end{enumerate}
\section{Data Collection Procedure}\label{sec:data_collection}
We similarly adopt the data collection procedure of \citet{clark-etal-2020-tydi}, which consists of question elicitation, article retrieval, and answer labeling.

\textbf{Question Elicitation} Our goal is to elicit \textit{information-seeking} questions, where the annotator asking the question is actually interested in the answer. To this end, annotators are shown short prompts consisting of the first 150 characters from Wikipedia articles in their language variety and asked to write questions that are not directly answered by the prompt and that the annotator is curious about. The intent behind the prompts is to inspire the raters to inquire about various topics; note however that we do not require that the generated questions actually pertain to the provided prompts. \citet{clark-etal-2020-tydi} provide a detailed justification for why it is important to elicit questions not directly answered by the prompts; briefly, this avoids questions that can be answered by simple techniques, resulting in a more challenging task. 

\textbf{Article Retrieval} For each elicited question, we leverage a Google search restricted to the Wikipedia domain for its respective language variety\footnote{For all Arabic varieties, only the questions are in that variety, while the articles are written in Modern Standard Arabic (see Section~\ref{sec:language_varieties}).} and retrieve the first result if any are found. If not found, the question is discarded and not included in our dataset. Note that we process the retrieved text to remove tables, long lists, and infoboxes, so that our resulting dataset is focused on natural text.

\textbf{Answer Labeling} Each found article is paired with its corresponding question and presented to 1 or 3 human annotators: 1 for our train split, 3 for our development and test splits. After first confirming that the question follows the guidelines discussed above, the annotator selects a paragraph in the article that contains the answer, or indicates that there is no such paragraph. The annotator then selects a \textbf{minimal answer}: a span of characters in the paragraph that is as short as possible while still comprising a satisfactory answer to the question. For example, for the question ``Who was the first President of the United States?'', annotators should select ``George Washington'' as the minimal answer. While many questions have a minimal answer of just a few words, others (such as ``What is an atom?'') require most of a sentence to answer.

We included multiple quality control and training steps in our data collection pipeline. For each rater, we first elicited a small number of English questions and provided feedback on any that did not meet our guidelines. Then we elicited a small number of in-language questions and had them reviewed by trusted in-house native speakers to verify that each rater is a native speaker of their respective language variety. In the answer labeling phase, we first had our raters review the task instructions and take a multiple-choice quiz covering how to annotate 21 example English question/article pairs, alternating between taking the quiz and reviewing the instructions until they achieved a score greater than 90\%. After passing, raters performed answer labeling on a small number of English examples which we manually reviewed and provided feedback on; raters completed this stage repeatedly (with different examples each time) until we were satisfied with their task understanding. Our raters are paid fair market rates for their time, including training time.
\section{Related Work}\label{sec:related_work}
Many existing extractive question answering datasets evaluate \textit{reading comprehension}, including SQuAD 2.0 \citep{rajpurkar2018know}, XQuAD \citep{artetxe2019cross}, MLQA \citep{lewis2019mlqa}, M2QA \citep{englander2024m2qa}, and LAReQA \citep{roy2020lareqa}. The common feature of these datasets is that each question was written after reading the paired passage. We contrast this with \textit{information-seeking} QA datasets, as in this work, where the question was written independently of the passage via a process designed to elicit questions that the question writer is curious about; examples include Natural Questions \citep{kwiatkowski2019natural} and its multilingual counterpart MKQA \citep{longpre2021mkqa}, as well as TyDi QA \citep{clark-etal-2020-tydi}.

Our dataset includes questions written in Arabic regional varieties with passages in Modern Standard Arabic. This is related to cross-lingual QA datasets such as XOR TyDi QA \citep{asai2020xor}, XTREME-UP \citep{ruder2023xtreme}, XQuAD \citep{artetxe2019cross}, MLQA \citep{lewis2019mlqa}, and LAReQA \citep{roy2020lareqa}, though none of those evaluate cross-variety QA. Conversely, existing datasets examining Arabic dialects such as MSDA \citep{boujou2021open}, IADD \citep{zahir2022iadd}, and MADAR \citep{bouamor2019madar} are not QA datasets.

Our dataset includes Tajik and Azerbaijani, two low-resource language varieties. Examples of existing work in these varieties include \citet{dovudov2012pos}, \citet{hecking2010tajik}, \citet{merchant2024parstext}, and \citet{isbarov2024open}. None of these provided QA datasets for these varieties.

Our dataset emphasizes long-context modeling by requiring models to search for an answer within an entire article (see Table~\ref{tab:data_stats} for statistics on average article length). We expect this to be useful to evaluate the capabilities of models from recent years designed to process text of this length \citep{beltagy2020longformer, zaheer2020big, ainslie2020etc, liu2024lost, reid2024gemini}.

One motivation for our decision to elicit questions in-language instead of creating a parallel QA dataset by translating a fixed set of questions is to ensure that the resulting questions are culturally relevant within each language. This motivation is shared by CaLMQA \citep{arora2024calmqa}, a QA dataset designed to be culturally relevant.

Our primary focus in this work is on developing an extractive QA dataset that: (1) contains information-seeking questions; (2) includes a set of languages that balances diversity (to assess coverage) and relatedness (to facilitate transfer learning); and (3) requires long-context capabilities. To our knowledge, ours is the first dataset to cover all of these criteria. 
\section{Selected Language Varieties}\label{sec:language_varieties}

This section provides a brief description of each language variety included in our dataset, grouped by language family (and branches in some cases). IETF BCP 47 language tags are provided for all varieties.

The 10 language varieties in our dataset cover 3 different  language families. Including multiple language families helps to assess models' language coverage, while including multiple related languages from within the same family allows practitioners using our dataset to evaluate transfer learning.

\subsection{Afro-Asiatic Language Family}
6 varieties from this language family are included.

\subsubsection{Arabic}
The \textit{questions} in our dataset include 4 varieties of Arabic, while the \textit{articles} are all from Arabic Wikipedia which is written in a fifth variety (Modern Standard Arabic). All varieties use the Arabic alphabet.

\textbf{Modern Standard Arabic (\textit{ar})} is primarily a literary variety of Arabic, and not generally spoken as a first language.

\textbf{Algerian Arabic (\textit{arq})} is a variety of Arabic spoken primarily in Algeria.

\textbf{Egyptian Arabic (\textit{arz})} is a variety of Arabic spoken primarily in Egypt. It is the most widely-spoken Arabic variety.

\textbf{Iraqi Arabic (\textit{acm})} is a variety of Arabic spoken primarily in Iraq, Syria, Turkey, Iran, and Kuwait.

\textbf{Jordanian Arabic (\textit{apc-JO})} is a variety of Arabic spoken primarily in Jordan.

\subsubsection{Northwest Semitic}
\textbf{Hebrew (\textit{he})} is primarily spoken in Israel and is written in the Hebrew script.
\subsection{Indo-European Language Family}

\textbf{Armenian (\textit{hy})} is spoken primarily in Armenia and written in the Armenian alphabet. It is the only member of an independent branch of the Indo-European language family.

\subsubsection{Persian}
We include two mutually-intelligible Persian language varieties.

\textbf{Farsi (\textit{fa})}, also known as Iranian Persian or just Persian, is spoken primarily in Iran and written in the Persian alphabet.

\textbf{Tajik (\textit{tg})}, also known as Tajiki Persian or just Tajiki, is spoken primarily in Tajikistan and written in the Cyrillic alphabet.

\subsection{Turkic Language Family}
Our dataset includes two mutually-intelligible language varieties from the Turkic family.

\textbf{Azerbaijani (\textit{az})} is a variety spoken primarily in Azerbaijan. In Azerbaijan it is written in Latin script, though elsewhere it is sometimes written in Arabic, Cyrillic, or Georgian scripts. In our dataset, questions and articles are in Latin script.

\textbf{Turkish (\textit{tr})} is a variety spoken primarily in T{\"u}rkiye (also known as Turkey) and written in Latin script.
\section{Dataset Statistics and Examples}\label{sec:dataset_statistics}

Our dataset contains a total of 28,197 questions, including 16,200 examples in the training split (with 1 annotation each), 5,995 dev examples (3 annotations each), and 6,002 test examples (3 annotations each). Table~\ref{tab:data_stats} lists basic statistics about our dataset, including the number of examples by language and split. It also lists the percentage of examples where a majority of annotations were \textsc{NULL}, meaning that the annotator did not find an answer in the article (see Section~\ref{sec:evaluation} for further discussion on treatment of \textsc{NULL} annotations). In many language varieties, the proportion of NULL examples is quite high. We believe that this is primarily due to these varieties' Wikipedias being comparatively small: the smaller number of articles makes it more likely to match a question with an irrelevant article, and the shorter average length of articles makes it less likely for the answer to be present.

Some examples from our dataset are shown in Figure~\ref{fig:data_examples}. In the first Turkish example, two annotators agreed exactly on the answer while a third did not find an answer (\textsc{NULL)}. In this case, the \textsc{NULL} annotation will be discarded in score calculation (see Section~\ref{sec:null_consensus}). In the second Turkish example, all 3 annotators agreed that the yes-no question's answer was \textsc{YES}, which is supported by the article (a relevant excerpt has been selected for the figure). In the first Azerbaijani example, there is disagreement among the annotators about whether to include the day (July 13) in the example alongside the year (1318). This illustrates that training raters to annotate in exactly the same way is difficult. In the second Azerbaijani example, the article does not contain the answer, but one annotator selected a plausible yet incorrect answer. This is an example of our use of 3 annotators for the evaluation set, along with the \textsc{NULL} consensus procedure, mitigating noisy annotations.

\begin{figure*}
    \centering
    \begin{examplebox}[Turkish]
    \textbf{Question:} \textit{Kaç tür asit vardır?} (How many types of acids are there?)
    
    \textbf{Article:} [...] \textit{Asitler başlıca iki grupta toplanabilirler:} [...] (Acids can be divided into two main groups:)

    \textbf{Answer Annotations:}
    \begin{enumerate}[topsep=0pt,itemsep=-1ex]
        \item \textit{iki} (two)
        \item \textit{iki} (two) 
        \item \textsc{NULL}
    \end{enumerate}
    \end{examplebox}
    \begin{examplebox}[Turkish]
    \textbf{Question:} \textit{rRNA katlanma yapar mı?} (Does rRNA fold?)
    
    \textbf{Article:} [...] \textit{Ribozomdaki proteinler rRNA\'nın belli kısımlarını tanıyıp oralara bağlanır, ardından bu proteinler arasındaki etkileşimler rRNA'nın daha da katlanmasına neden olur ve sonunda ribozom meydana gelir.} [...] (Proteins in the ribosome recognize and bind to certain parts of the rRNA, then the interactions between these proteins cause the rRNA to fold further, eventually forming the ribosome.)

    \textbf{Answer Annotations:}
    \begin{enumerate}[topsep=0pt,itemsep=-1ex]
        \item \textsc{YES}
        \item \textsc{YES}
        \item \textsc{YES}
    \end{enumerate}
        
    \end{examplebox}
    
    \begin{examplebox}[Azerbaijani]
    \textbf{Question:} \textit{Fəzullah Rəşidəddin nə zaman vəfat edib?} (When did Fazullah Rashiduddin die?)
    
    \textbf{Article:} [...] \textit{O, Elxani hökmdarı Olcaytunu zəhərləməkdə mühakimə olunaraq 13 iyul 1318-ci ildə, 70 yaşında edam edilmişdir.} [...] (He was tried for poisoning the Elkhanid ruler Oljait and executed on July 13, 1318, at the age of 70.)

    \textbf{Answer Annotations:}
    \begin{enumerate}[topsep=0pt,itemsep=-1ex]
        \item \textit{1318} (1318)
        \item \textit{1318} (1318) 
        \item \textit{13 iyul 1318} (July 13, 1318)
    \end{enumerate}
    \end{examplebox}
    \begin{examplebox}[Azerbaijani]
    \textbf{Question:} \textit{Ukraynada ilk konsitusiya nə vaxt qəbul olunub?} (When was the first constitution adopted in Ukraine?)
    
    \textbf{Article:} [...] \textit{28 iyun — Ukraynanın Konstitusiya günü.} [...] (June 28 — Constitution Day of Ukraine.)

    \textbf{Answer Annotations:}
    \begin{enumerate}[topsep=0pt,itemsep=-1ex]
        \item \textsc{NULL}
        \item \textsc{NULL}
        \item \textit{28 iyun} (June 28)
    \end{enumerate}
    \end{examplebox}
    \caption{Examples from our development set. The articles have been trimmed to the relevant portion for clarity. Minimal span answers are represented as byte ranges in our dataset, but in this figure, the text indicated by those ranges is shown. Note that the English glosses are not included in the dataset.}
    \label{fig:data_examples}
\end{figure*}
  
\begin{table*}
\centering
\begin{tabular}{ c|c|c|c|c|c|c|c }
Language & Train & Dev & Test & Avg. & Avg. & Avg. & \% With \\
Variety & Examples & Examples & Examples & Question & Article & Answer & \textsc{NULL} \\
 & (1 Rating) & (3 Ratings) & (3 Ratings) & Tokens & Bytes & Bytes & Consensus \\
\hline
Algerian Arabic & 1306 & 649 & 647 & 5.8 & 92K & 145 & 69.6\% \\
Egyptian Arabic & 1519 & 754 & 755 & 6.3 & 131K & 60 & 49.2\% \\
Jordanian Arabic& 1386 & 688 & 690 & 5.3 & 114K & 84 & 50.7\% \\
Iraqi Arabic & 1320 & 658 & 657 & 6.6 & 91K & 42 & 66.2\% \\
Armenian & 1673 & 823 & 824 & 7.0 & 119K & 39 & 76.7\% \\
Azerbaijani & 1469 & 724 & 724 & 6.0 & 48K & 18 & 68.1\% \\
Hebrew & 1517 & 760 & 759 & 6.4 & 75K & 43 & 48.8\% \\
Farsi & 3275 & 105 & 104 & 8.1 & 73K & 46 & 47.9\% \\
Tajik & 1114 & 86 & 88 & 5.9 & 37K & 36 & 82.7\% \\
Turkish & 1624 & 749 & 754 & 5.2 & 36K & 30 & 51.4\% \\

\end{tabular}
\caption{Basic data statistics. ``Avg. Question Tokens'' is the average number of whitespace-delimited tokens in questions. ``Avg. Answer Bytes'' is the average number of bytes of all minimal answers where the consensus is non-\textsc{NULL}. ``\% With \textsc{NULL} Consensus'' is the percentage of examples where the consensus is \textsc{NULL}. These values are calculated over all examples from all splits.}
\label{tab:data_stats}
\end{table*}

\section{Evaluation}\label{sec:evaluation}

The primary metric for our dataset is average per-example F1, though we also report exact match (EM) scores. Both F1 and EM for each example are calculated as the maximum score against each of the three available annotations for that example, with one large exception related to the distinction between \textsc{NULL} and non-\textsc{NULL} answers. Details on this distinction, as well as the specifics of score calculation, are presented in this section.

\subsection{\textsc{NULL} Consensus}\label{sec:null_consensus}
For many questions in our dataset, some or all annotators were unable to find an answer in the presented article. This is especially true for some language varieties where the associated Wikipedia is relatively small. As argued in \citet{clark-etal-2020-tydi}, model performance may be artificially inflated if a model could get credit for predicting \textsc{NULL} when \textbf{any} annotation is \textsc{NULL}. Thus, we first establish a ``\textsc{NULL} consensus'' for each example in our evaluation splits: if fewer than two annotations are \textsc{NULL}, then any \textsc{NULL} annotations are discarded. Conversely, if at least two are \textsc{NULL}, then any non-\textsc{NULL} annotations are discarded. This essentially means that we take a majority vote to determine whether there are any valid answers to the question, which is intended to limit noise in the final labels. For the training split, where only one annotation is present, the consensus is \textsc{NULL} if and only if the provided annotation is \textsc{NULL}.

\subsection{F1 Score Calculation}
For each example, if the model produces \textsc{YES}, \textsc{NO}, or \textsc{NULL}, then the F1 score for that example is $1.0$ if any non-discarded annotation (see Section~\ref{sec:null_consensus}) matched that answer, and $0.0$ otherwise. If the model produces valid start and end byte indices (referring to a candidate minimal span answer within the article), the indices in that range are treated as a set and compared to the set of indices selected in each annotation that contains a minimal answer span: the F1 score against a single annotation is the harmonic mean of precision and recall over byte indices, and the score for the example is the maximum score over available annotations ($0.0$ if no annotations contained a minimal span answer), to account for annotator variability in selecting the minimal span. The per-example F1 scores are then averaged to produce the model's final score.

\subsection{Exact Match Calculation}
The EM score is the proportion of examples where the model's output exactly matched an available annotation. It is calculated in exactly the same way as F1 except when comparing minimal span answers between the model and annotations: instead of assigning partial credit, the score is $1.0$ if the byte ranges exactly match, and $0.0$ otherwise. Thus, the EM score is never higher than the F1 score.

\section{Baseline Systems}\label{sec:baseline_systems}
To provide baseline results as a point of comparison for future work on our dataset, we present results from two publicly-available models on our task: Gemini 1.5 Pro \cite{geminiteam2024gemini15unlockingmultimodal} and Gemini 2.0 Flash\footnote{https://blog.google/technology/google-deepmind/google-gemini-ai-update-december-2024/}. Both are large language models trained on multilingual text.

The input consists of three components: 1) The preamble, 2) one exemplar of each question type for in-context learning, and 3) the input question. For each language, a set of exemplars was collected and used for all samples within that language.

\begin{samepage}
The following is the preamble used:
\begin{quote}
\textit{Consider the following question and passage, which may contain an answer. Find the answer, if it exists. If there is not a good and complete answer to the question, respond with "no answer". If the question is asking for a yes or no answer, return "yes" or "no" if there is evidence for this in the passage.  Answers will usually be just a few words, but could be more in some cases.}
\end{quote}
\end{samepage}

Each exemplar is formatted as follows: \newline
\begin{quote}
Article:\textit{<article text>}\newline
Q:\textit{<question text>} \newline
A:\textit{<answer text>} \newline
\end{quote}
For yes/no answers, the answer text is ``yes'' or ``no'', and for \textsc{NULL} answers, the answer text is ``no answer''; evaluation is case-insensitive for these answer types. For minimal span answers, the task definition requires a byte range but off-the-shelf LLMs cannot reliably produce them, so instead the model produces the answer text directly and, as a heuristic, we locate the first occurrence of that text within the article to recover the byte range.

Different exemplars are concatenated by a newline. In order to reduce the total input length to the model, we selected exemplars from the shorter articles.
We use greedy sampling with the maximum generation length of 1024.

\subsection{No-Answer Critic}

While we direct the baseline models to output ``no answer'' when it cannot find an answer to the question, we observed that they sometimes produced responses indicating that no answer was found but in the wrong format, such as ``I can't find an answer to this question'' (either in English or in the question's language variety). Because we are not primarily interested in measuring the models' adherence to this particular output format,  we use a second inference pass with the model to analyze its own output to determine whether the answer is a misformatted version of ``no answer''. We only use this no-answer critic when the model's output is not a valid substring of the article, and also none of ``yes'', ``no'', or ``no answer'' (case-insensitive). We observed this critic to be very effective at detecting \textsc{NULL} cases that did not follow the requested output format. See Appendix~\ref{app:sec:no_answer} for the prompt used for the no-answer critic.

\section{Results and Discussion}\label{sec:results}

\begin{table}
\centering
\begin{tabular}{ c|c|c|c|c }
 Gemini Model $\rightarrow$ & \multicolumn{2}{c|}{1.5 Pro} & \multicolumn{2}{c}{2.0 Flash}\\
 Language $\downarrow$ & EM & F1 & EM & F1\\
 \hline
Algerian Arabic	&	46.5	&	48.2	&	47.3	&	48.9	\\
Egyptian Arabic	&	54.1	&	57.9	&	54.9	&	58.6	\\
Iraqi Arabic	&	55.3	&	56.9	&	56.2	&	58.0	\\
Jordanian Arabic	&	46.9	&	54.6	&	48.8	&	56.4	\\
Armenian	&	74.0	&	74.8	&	72.7	&	73.6	\\
Azerbaijani	&	71.8	&	74.4	&	71.5	&	74.3	\\
Hebrew	&	50.0	&	52.8	&	50.5	&	52.7	\\
Farsi	&	38.1	&	41.9	&	40.0	&	48.3	\\
Tajik	&	73.3	&	74.2	&	66.3	&	67.1	\\
Turkish	&	55.7	&	58.0	&	56.3	&	58.4	\\
\end{tabular}
\caption{Exact Match and F1 scores (both out of 100) of our baseline systems on all 10 language varieties on our development set.}
\label{tab:dev_results}
\end{table}

\begin{table}
\centering
\begin{tabular}{ c|c|c|c|c }
Gemini Model $\rightarrow$ & \multicolumn{2}{c|}{1.5 Pro} & \multicolumn{2}{c}{2.0 Flash}\\
 Language & EM & F1 & EM & F1 \\
 \hline
Algerian Arabic	&	45.0	&	46.9	&	45.7	&	47.7	\\
Egyptian Arabic	&	53.0	&	58.4	&	53.1	&	58.0	\\
Iraqi Arabic	&	60.6	&	62.9	&	61.5	&	64.3	\\
Jordanian Arabic	&	46.8	&	56.3	&	48.1	&	57.6	\\
Armenian	&	71.8	&	72.7	&	71.1	&	72.2	\\
Azerbaijani	&	74.9	&	76.2	&	73.6	&	75.1	\\
Hebrew	&	53.4	&	57.3	&	52.6	&	56.3	\\
Farsi	&	34.6	&	38.7	&	45.2	&	50.4	\\
Tajik	&	68.2	&	68.2	&	56.8	&	57.6	\\
Turkish	&	53.3	&	55.1	&	53.8	&	55.3	\\
\end{tabular}
\caption{Exact Match and F1 scores (both out of 100) of our baseline system on all 10 language varieties on our test set.}
\label{tab:test_results}
\end{table}

Tables~\ref{tab:dev_results} and \ref{tab:test_results} report our baseline's performance on the development and test splits, respectively. Neither baseline clearly outperforms the other overall. For both models, scores vary widely between language varieties. We note that this is not sufficient to conclude that a model is better at answering questions in e.g. Armenian than in Farsi: the questions are different, and specifically the proportion of \textsc{NULL}-consensus answers in Armenian is higher than in Farsi. We therefore recommend that practitioners using our dataset focus on comparing the scores of different models within the same language varieties to measure per-variety improvement, as opposed to comparing scores across varieties for the same model.

By comparing EM and F1 scores, we can estimate the frequency with which the baseline models predict an answer span that overlaps with, but does not exactly match, a gold answer span. The fact that EM and F1 scores are relatively close to each other in Tables~\ref{tab:dev_results} and \ref{tab:test_results} indicates that this phenomenon is fairly uncommon for these baseline models.

\begin{table}[t]
    \centering
    \begin{tabular}{c|c|c|c|c}
    Gemini Model $\rightarrow$ & \multicolumn{2}{c|}{1.5 Pro} & \multicolumn{2}{c}{2.0  Flash}\\
    \textsc{NULL} consensus $\rightarrow$ & \Checkmark & \XSolidBrush & \Checkmark & \XSolidBrush\\
    Language $\downarrow$ & & & & \\
         & & & & \\
         \hline
Algerian Arabic	&	57.0	&	16.9	&	57.8	&	17.2	\\
Egyptian Arabic	&	63.2	&	53.1	&	64.3	&	53.3	\\
Iraqi Arabic	&	60.3	&	48.8	&	61.6	&	49.4	\\
Jordanian Arabic	&	58.7	&	50.2	&	61.5	&	51.0	\\
Armenian	&	85.0	&	39.5	&	83.7	&	38.7	\\
Azerbaijani	&	82.7	&	58.2	&	81.7	&	59.8	\\
Hebrew	&	60.1	&	45.3	&	61.9	&	43.2	\\
Farsi	&	20.0	&	45.5	&	6.7	&	55.2	\\
Tajik	&	75.7	&	65.4	&	67.6	&	64.0	\\
Turkish	&	61.6	&	54.2	&	61.9	&	54.7	\\
    \end{tabular}
    \caption{Baseline F1 results on our development set, divided into examples with a \textsc{NULL} (\Checkmark) vs. non-\textsc{NULL} (\XSolidBrush) consensus.}
    \label{tab:null_results_breakdown_dev}
\end{table}

\begin{table}[t]
    \centering
    \begin{tabular}{c|c|c|c|c}
    Gemini Model $\rightarrow$ & \multicolumn{2}{c|}{1.5 Pro} & \multicolumn{2}{c}{2.0  Flash}\\
    \textsc{NULL} consensus $\rightarrow$ & \Checkmark & \XSolidBrush & \Checkmark & \XSolidBrush\\
    Language $\downarrow$ & & & & \\
         & & & & \\
         \hline
Algerian Arabic	&	55.9	&	16.1	&	56.9	&	16.3	\\
Egyptian Arabic	&	62.8	&	54.6	&	63.3	&	53.5	\\
Iraqi Arabic	&	66.7	&	54.4	&	68.0	&	56.2	\\
Jordanian Arabic	&	62.2	&	50.4	&	64.2	&	51.0	\\
Armenian	&	81.6	&	40.3	&	81.1	&	39.4	\\
Azerbaijani	&	83.0	&	60.9	&	81.4	&	61.0	\\
Hebrew	&	64.6	&	50.3	&	63.5	&	49.4	\\
Farsi	&	46.7	&	37.4	&	46.7	&	51.0	\\
Tajik	&	71.9	&	58.3	&	62.5	&	44.6	\\
Turkish	&	56.0	&	54.1	&	56.8	&	53.6	\\
    \end{tabular}
    \caption{Baseline F1 results on our test set, divided into examples with a \textsc{NULL} (\Checkmark) vs. non-\textsc{NULL} (\XSolidBrush) consensus.}
    \label{tab:null_results_breakdown_test}
\end{table}

Given the high proportion of \textsc{NULL}-consensus examples in some languages, one concern is that a na{\"i}ve baseline could simply predict ``no answer'' in all cases and achieve reasonable performance. It is therefore important to understand performance on the non-\textsc{NULL} portion as well. Tables~\ref{tab:null_results_breakdown_dev} and \ref{tab:null_results_breakdown_test} present this breakdown, using 100-scaled F1 score\footnote{For the \textsc{NULL} portion, F1 is equivalent to exact match.}. Both baseline models achieve a higher performance on the \textsc{NULL} portion of both development and test splits in all languages except for Farsi. However, with the possible exception of Algerian Arabic, performance on the non-\textsc{NULL} data is still reasonable, indicating that the overall performance reported in Tables~\ref{tab:dev_results} and \ref{tab:test_results} is not a result of always predicting ``no answer''. We recommend that practitioners benchmarking performance on our dataset also report scores broken down in this way.

We emphasize that these baselines were not finetuned for this task, and that inference consists of showing the entire article to the model at once. Publicly-available models capable of achieving reasonable performance on such a long-context task are a recent development, and make it much easier for researchers to experiment on our task. As a point of contrast, the baseline for the original TyDi QA dataset \citep{clark-etal-2020-tydi} used a finetuned model that scored each candidate short answer span separately. We believe that the technological advances since then will make the MinSpan task easier for researchers to iterate on.

\section{Conclusion}
In this work we have presented a dataset of information-seeking questions in 10 low-resource language varieties, and demonstrated the feasibility of using modern LLMs to extract answers to these questions from large text contexts. By releasing the data and code to use it, we hope to facilitate the measurement and improvement of models' performance in these language varieties.

\bibliography{tacl2021}
\bibliographystyle{acl_natbib}

\appendix
\section{Appendix}
\subsection{No-Answer Critic Prompt}\label{app:sec:no_answer}
Our no-answer critic used the following prompt:
\begin{quote}
\textit{You are given a text and you should check whether it indicates that an answer was not found or there is no answer, answer with no or yes only. Here are some examples.\newline
                     text: 25 AA response: No \newline 
                     text: here AA response: No \newline
                     text: 2020 AA response: No \newline
                     text: YES AA response: No \newline
                     text: No AA response: No \newline
                     text: No Answer AA response:Yes \newline
                     text: there is no information in the text to answer the question AA response: Yes \newline
                     text: he was born in germany AA response: No}
\end{quote}

\textit{AA} is an arbitrary string separating the text and critic response. Other formatting can be used as well.

\end{document}